\documentclass{article}

\usepackage{arxiv}

\usepackage[utf8]{inputenc} 
\usepackage[T1]{fontenc}    
\usepackage{hyperref}       
\usepackage{url}            
\usepackage{booktabs}       
\usepackage{amsfonts}       
\usepackage{nicefrac}       
\usepackage{microtype}      
\usepackage{lipsum}
\usepackage{graphicx}

\title{Development and verification of a simulation for leveraging results of a human subjects programming experiment}

\author{
  Achim Gerstenberg \\
  Department of Mechanical and Industrial Engineering\\
  Norwegian University of Science and Technology\\
  Trondheim, Norway \\
  \texttt{achim.gerstenberg@ntnu.no} \\
   \And
 Martin Steinert \\
  Department of Mechanical and Industrial Engineering\\
  Norwegian University of Science and Technology\\
  Trondheim, Norway \\
   \texttt{martin.steinert@ntnu.no} \\
}

\begin{document}
\maketitle

\begin{abstract}
Quantitatively evaluating and comparing the performance of robotic solutions that are designed to work under a variety of conditions is inherently challenging because they need to be evaluated under numerous precisely repeatable conditions  Manually acquiring this data is time consuming and imprecise. A deterministic simulation can reproduce the conditions and can evaluate the solutions autonomously, faster and statistically significantly. We developed such a simulation designated to leverage data from a human-subject experiment post-experimentally. We present the development of the simulation and the verification that it actually reproduces the results obtained with the physical robot.
The aim of this publication is to provide insight into the development details such that other researchers can replicate the setup and to show the degree of validity of the simulation.
\end{abstract}


\section{Introduction}
Gerstenberg and Steinert \cite{Gerstenberg_2018_a} presented an experimental setup for engineering design methodology research where participants are expected to program a mobile robot to autonomously detect and move objects in the shortest time possible and from any starting position within a given area. The time the robot needs to complete the task is the performance measure that can be used for testing the influence of different design methodologies on performance outcome in such an open-ended task. During the experiment the participants generate and test their solutions. The test results during the experiment are not sufficient to obtain statistically valid performance results but the codes that the participants generate can be evaluated post-experimentally.
This can be done by loading the participant codes onto the robot and then manually executing the codes, observing the robot and documenting its performance from many different starting positions. While gathering this data the conditions must be kept as equal as possible for comparability. This means that the starting conditions must be accurately repeated, the battery equally charged, frictions in the motors remain equal, etc.. To reduce the error introduced by not perfectly replicating the conditions, each starting position can be repeatedly executed and the results for this starting position are then averaged. Acquiring data manually from many starting position is very time consuming and annoying.
Therefore, this paper presents a digital and deterministic simulation that speeds up the post-experimental performance evaluation and, in addition, makes it more comparable by guaranteeing equal conditions when comparing solutions. 
We describe the task and the robot used in the experiment that this simulation is made for, how we developed the simulation from observing the physical robot, how we validate that the simulation gives qualitatively and quantitatively valid results and use the simulation to illustrate that small perturbations in the conditions can have a significant effect on performance when evaluating only a single or few repetitions.

\section{The physical robot and the task}
The robot is built from LEGO technic and is controlled by a LEGO Mindstorms NXT 2.0 system. Two electric motors can move the robot via two belts; one on the left and one on the right side. The robot has in total 4 sensors. A ultrasound distance sensor is mounted in the center front and two identical light/colour sensors are mounted on the sides. The light/colour sensor can detect light intensity and measure the colour of nearby objects. These three sensors point forward. Near the front and pointing downwards, the robot has a reflection sensor. The sensor emits red light and measures how much of this emitted light is reflected back into the sensor. The reflection sensor can detect differences in reflectivity of the surface underneath the robot. 
\par The robot is programmed by the participant in the NXC language. NXC stands for ``not exactly C'' and is a programming language very similar to C. The participant is provided with a library that includes functions specifically written for this robot. It simplifies the interpretation of raw data such as converting the time of flight of the ultrasound pulse from the ultrasound distance sensor to a distance in centimeters or the raw light sensor values into a meaningful value for detecting blinking lights. The library allows the participant to quickly use the robot to solve the task instead of spending time on programming the basic robot functionality.

A rectangular cardboard playground platform of approximately 1,50 m by 1,20 m has a white area rectangle in the centre surrounded by a 17 cm cardboard fringe. On top of the white area three coloured cubes (red, green and blue) are placed. The participant programs the robot so that the robot autonomously removes the cubes entirely from the white area in the shortest possible time.
In the top of the cube a cut-out allows inserting one blinking light per cube that can be detected by the robot from any direction.
The participants can optionally place up to three blinking lights anywhere on the playground including inside the top of the cube. The blinking lights are detectable with the light/colour sensor and using the blink variable from the library. The robot can detect the difference of reflectivity between the cardboard surface and the white area with the downwards pointing reflection sensor. If the robot drives off the cardboard playground the task is failed. The robot shall be capable to solve the task from any starting position and orientation inside the white area.

\section{Development approach}
In contrast to most simulations that are used to test different solutions before building the physical robot this simulation was developed after the physical robot and environment already existed in order to use the simulation to generate a performance measure in a controlled, repeatable and faster way. 
The approach for developing the simulation is a repeating cycle of measuring the physical robot behaviour and then digitally replicating it and comparing the real robot behaviour to the simulated robot behaviour. The movement of the physical robot and each sensor are measured individually and several times. These measurements are averaged. The averaged measurements describing the robot can be found in the datasheet~\cite{datasheet} and are in part shown in the next subsection. The physical environment and the robot behaviour were modelled in a 2D digital, scaled and deterministic simulation. For comparing the simulation with reality the simulation and the physical model are set up under similar conditions and each sensor and movement of the robot are compared individually. If the simulation is not accurate within the error of the measurements of the physical robot behaviour then a more detailed digital representation of the physical robot is developed and the comparison is repeated until the simulated robot behaviour is well within the error range of the measurements of the physical robot behaviour. If this cannot be achieved by making more and more detailed digital representations of the physical robot then we implement elements that can no longer be explained from the setup of the physical robot. We justify this approach because the aim is not to make a comprehensive explanatory model of how the physical robot behaves but the aim is to digitally replicate the physical robot's behaviour as precisely as possible to predict task performance outcome.

After calibrating the physical shapes and sizes, robot propulsion and sensors of the simulation individually, the simulation was compared to the behaviour of the physical robot in the context of solving the task. This comparison was done with three different codes that use as different task solving approaches as possible and cover all sensors and functions available to the participants. All three are tested from several different starting positions. The qualitative comparison is done by looking if the real and simulated trajectories are similar and the interaction of the robot and the cubes and edges of the white are phenomenologically similar. The quantitative comparisons are discussed in the verification section.

\section{Technical solution}
For displaying, animating, programming the robot behaviour and simulating the physical interactions between the objects we use the Unity game development software version 5.6.1 ~\cite{unity}.
The first step was to digitally replicate a top-down representation of the cardboard and the white area, the coloured cubes and the robot dimensions. This is done by drawing a scaled image of each of the objects and inserting them into the unity scene. All of these images have a box collider tracing the outline of the objects. Unity detects when two box colliders enter or exit into overlap and gives an event trigger that can be used for capturing the state of the robot in the environment.
The moving objects (robot and cubes) are modelled as a rigid body. This means that the physics engine of unity calculates the interacting forces between the objects and translates them accordingly. For simulating the deceleration of the cubes, the rigid body of the cubes is assigned a drag value. 
Onto the robot image/object the sensors are placed as images as well. Only the downwards reflection sensor uses a box collider. It is used to detect when this sensor is entering or exiting the white area or the cardboard area.

\subsection{code organization}
The unity scene animation is controlled by code written in C sharp. This includes the movement and the simulation of the sensors, coordinating the execution of the participant codes in the correct order from predefined starting conditions, documentation of the robot behaviour for later analysis, the simulation representation of the library that the participants used to control the physical robot and a reoccurring simulation loop that animates the scene and calculates the physics interactions between the objects. This calculation by the physics engine and the animation are executed during a code block called ``fixedUpdate()'' which is provided by the Unity game engine. FixedUpdate is executed 50 times a second. Additionally to the physics engine we use the fixedUpdate class to include our code for translating the robot in the scene, update the array that saves the sensor value for the blink sensor and check if the task is completed to reset it for the next simulation.
All other code evaluations are calculated in between two successive FixedUpdates. We use coroutines to coordinate the schedule for executing the participant codes. Coroutines are like threads that run between each FixedUpdate and can be yielded for specific times like until the next fixedUpdate or until another coroutine has finished. While FixedUpdate is like the relentless clock that runs until the simulation stops, coroutines can be used to execute code at specific times in a sequence. This, apart for scheduling the participant codes, is used to interrupt the execution of the participant code when those include the "wait" or "turn" command until the waiting time has elapsed or the robot has finished the turn.

\subsection{simulating motors}
\begin{figure}
  
  \centering
    \includegraphics[width=0.5\textwidth]{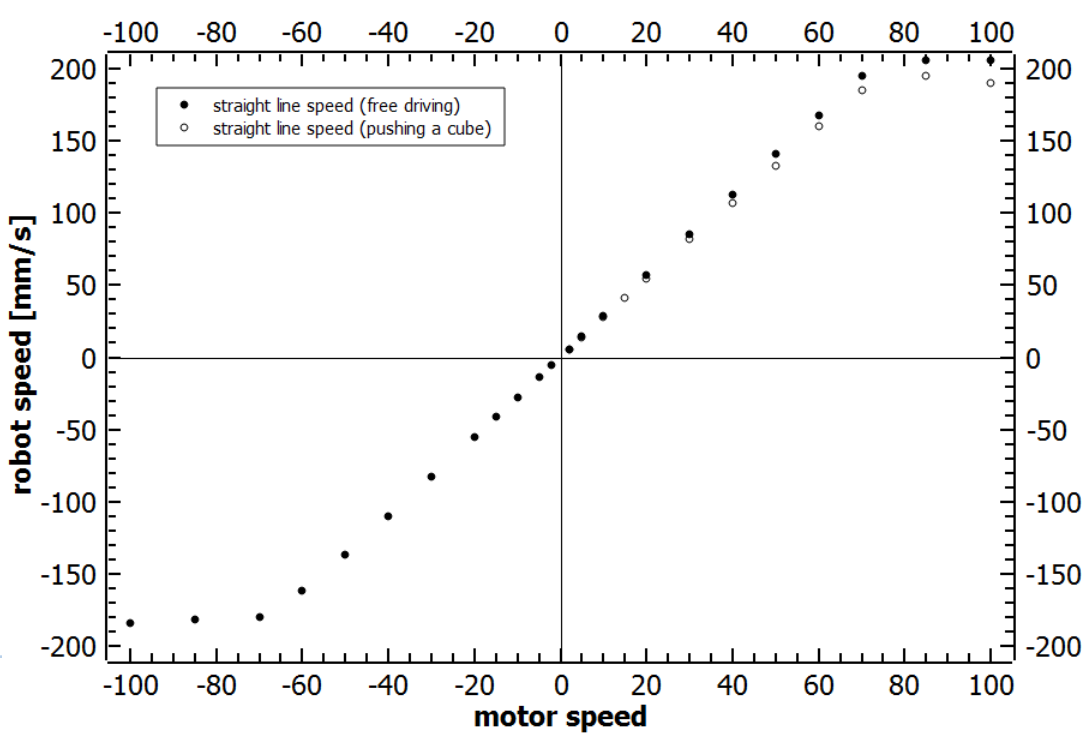}
    \caption{Motor speed as used in the participant code vs. physical distance driven per time for fully charged batteries, with and without pushing a cube.}
    \label{fig:speed}
\end{figure}

The movement of the digital robot is modelled with two speed state variables that determine the amount of translation of the robot object during the execution of FixedUpdate. They represent each of the driving belts of the physical robot. These two speed state variables are set by the motor and the turn function within the experiment library. The library is accessible to the participants while coding their solution. The translation along the forward axis of the robot is proportional to the average of those two speed variables and scaled by a calibration factor that is determined from the slope of a linear fit between -75\% and 75\% motor speed in figure \ref{fig:speed}. For motor speeds over 80\% the motors of the physical robot are not sufficiently strong to move the robot accordingly. Even with full  batteries the speed levels off at motor speeds higher than 75\% or below -75\%. This is modelled in the simulation by limiting the maximum speed to 75\% or -75\% respectively if a higher amount is written in the code. It is verified by comparing the times that the physical and simulated robot need at different coded motor speeds to cross the long side of the white area.
While pushing a cube the physical robot drives slower because the motors are not strong enough to fulfill the power demand required from the PID control. This occurs when pushing a cube with a desired motor speed above 20\%. For motor speeds above 20\% the robot speed is reduced by 5\% of the coded motor speed.

\begin{figure}
  
  \centering
    \includegraphics[width=0.5\textwidth]{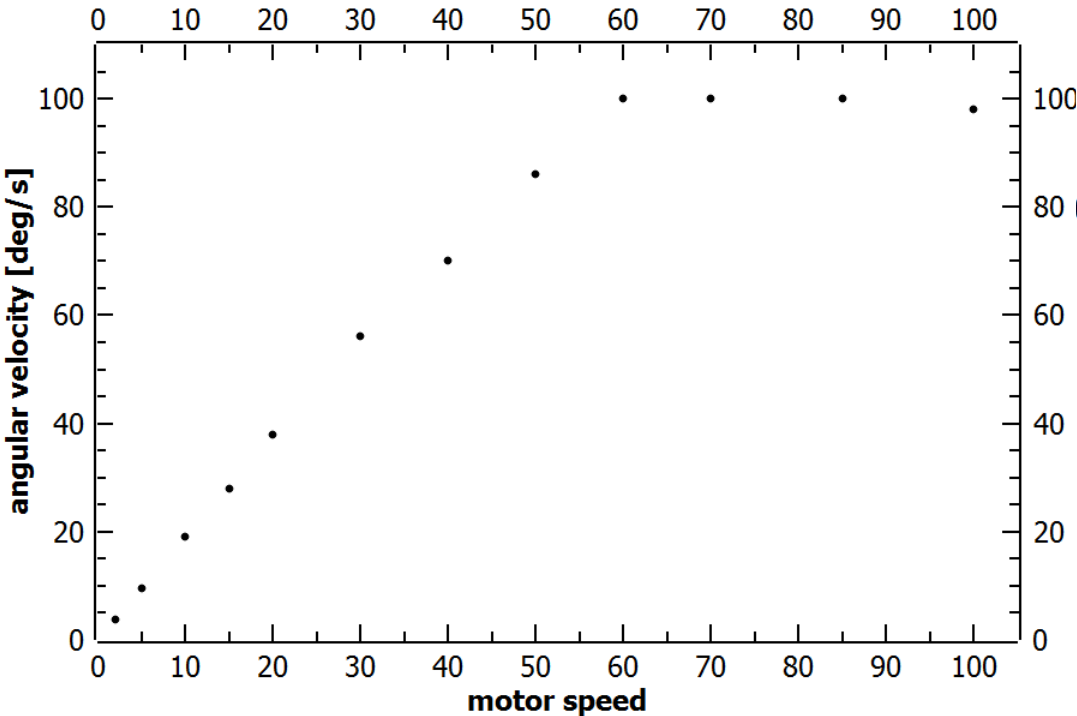}
    \caption{Angular velocity of the physical robot when having opposing speed state variables (rotating on the spot).}
    \label{fig:turnspeed}
\end{figure}

When the robot is driving in a turn, i.e. when the speed state variables for each side are not equal, then the simulated robot orientation is rotated proportional to the difference between the two speed state variables. In case the absolute value of the speed state variables are equal but they have opposite signs then the robot is turning on the spot. The proportionality factor is determined from the slope between 0\% and 60 \% motor speed of the angular velocity plot of the physical robot shown in figure \ref{fig:turnspeed} and limited above motor speeds of 60\%.

The motor speed in the simulation is programmed to change by a maximum of 9 percent points at the next FixedUpdate until it reaches the desired motor speed. In order to simulate the inertia of the physical robot and the acceleration limit was determined by qualitatively observing and comparing the behaviour of the physical and the simulated robot when changing from a rotation to a straight forward movement because it is difficult to accurately measure the acceleration of the physical robot otherwise.

Besides turning the robot on the spot by setting the two speed state variables, the library provided to the participant gives the possibility to turn the robot by a given amount of degrees and with a given rotation speed using the turn function from the library. In the simulation this is implemented using a coroutine that halts the participant code until the turn is completed. This means, similar to the physical robot, that the participant code cannot simultaneously execute another operation such as reacting to sensor inputs. In the simulation the turn is implemented by calculating the orientation where the turns need to stop and then calculating the target orientation for each FixedUpdate depending on the rotation speed. Then the turn coroutine sets the speed state variables to zero and the robot orientation is directly controlled by the coroutine and no longer by the code that is executed during FixedUpdate. However, the coroutine rotates the robot and is yielded after each rotation step until FixedUpdate has been executed to keep the turn synchronized with the physics simulation that occurs in FixedUpdate. 

\subsection{simulating sensors}

The units used during measuring the sensor outputs of the physical robot are centimeters for distances and degrees for angles. The angles in the physical world and the simulation are equivalent and need to translation. However, the centimeters measured in the real world do not correspond to the distance units used in the Unity editor. One distance unit in the Unity editor corresponds to $9.49223$ centimeter in the physical world. The measurement and fit functions shown below are in centimeters and the conversion number is used to translate from the real world to the simulated values. 

\subsubsection{Downwards reflection sensor}
The simulated downwards reflection sensor is a child object of the robot in the Unity scene. This means it uses the robot object as its reference frame and therefore moves synchronous with the robot. It has a box collider surrounding it. It is used to detect entering or exiting the box colliders of the white area or the cardboard area. Whenever the box collider of the reflection sensor enters or exits one of those area box colliders, a state variable is set to the according reflective value that was measured with the physical sensor. The reflectivity value for the white area is measured and simulated at 47, cardboard at 36 and outside the cardboard at 16.

\subsubsection{Ultrasound distance sensor}
\begin{figure}
  \centering
    \includegraphics[width=0.5\textwidth]{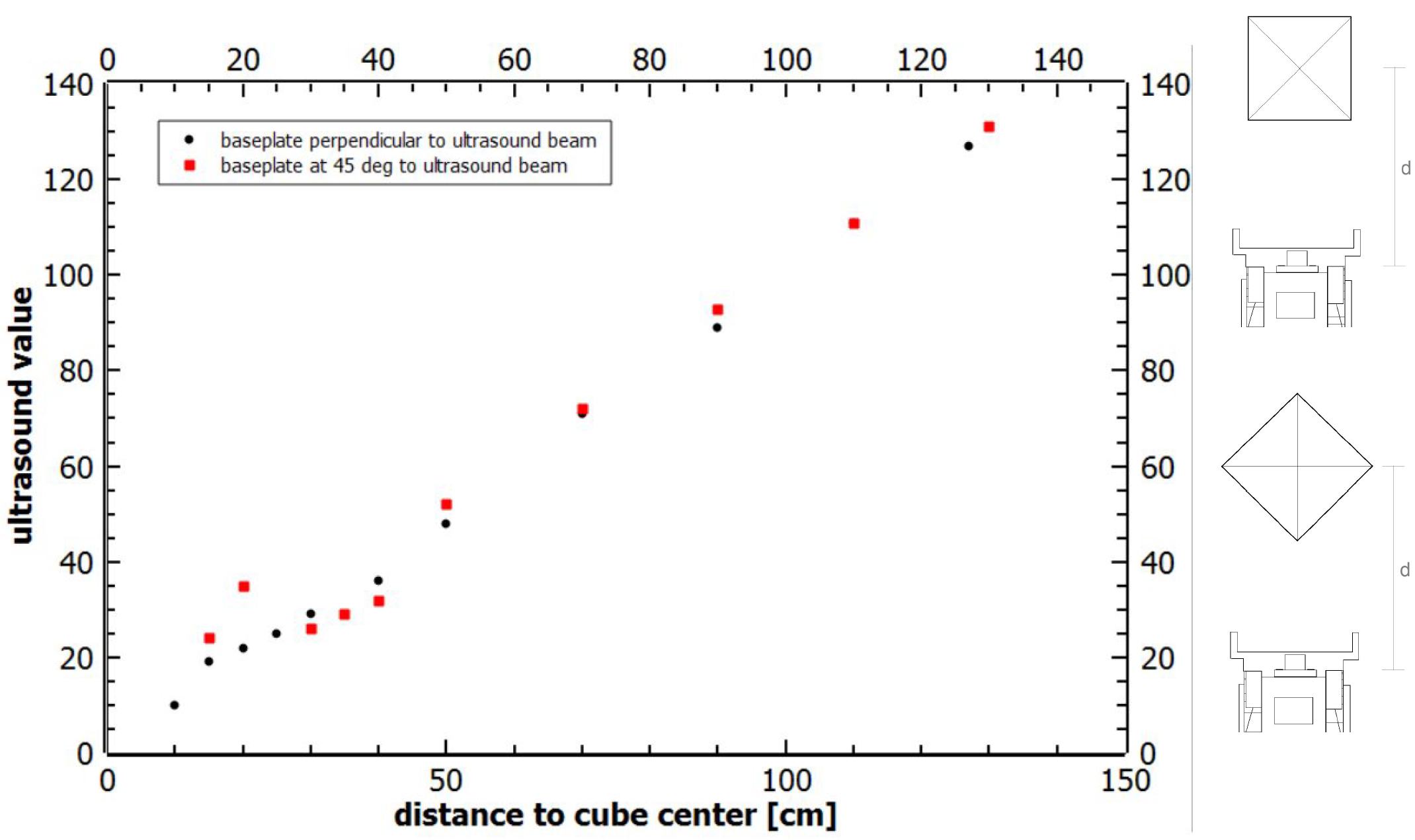}
    \caption{The plot shows the ultrasound value measured by the physical robot depending on the actual distance from the cube center. The sketches on the right side of the graph illustrate the cube orientation perpendicular to the ultrasound sensor (top) and at 45 degrees orientation (bottom).}
    \label{fig:ultrasound}
\end{figure}

The simulated ultrasound sensor is, similar to the downwards reflection sensor, also a child object of the robot in the Unity scene that moves with the robot object. The simulated sensor value is estimated from the distance between the center of the ultrasound sensor object and the center of the cube. The conversion between the simulation units and the centimeter used in the physical world is made with the slope of the measurement shown in figure \ref{fig:ultrasound} for when the cube base plate is orientated perpendicular to the ultrasound sensor. For angles where the cube's base plate is oriented at 45 degrees to the forward direction of the sensor, the distance measurement is similar but the cutoff angles - this is the largest angle where the sensor can still detect the cube - is smaller. For a perpendicular orientation the cutoff angle is $35 \pm 3$ degrees whereas at 45 degrees it is $22 \pm 2$. We model a linear decline of the cutoff angle according to these numbers: $cutoff = - \frac{13}{45} \cdot \gamma + 35$ where $\gamma$ is the angle between the forward direction of the ultrasound sensor and a vector perpendicular to the cube's base plate.

\subsubsection{Blink light sensor}

The blink value is determined from the forward pointing light / colour sensors on the left and right side of the robot. In the simulation those sensors are child objects of the robot object and move synchronous with the robot. 
The sensors on the physical robot detect the blink light by the light intensity difference between the times when the blink light is turned on and turned off. This requires time for the blink light to turn on and off again and causes a delay in the blink light measurement. In the simulation this is modelled by assessing delayed values from an array that stores the past blink values. In this simulation we use a delay of 500 milliseconds. The simulation, as well as the physical robot, evaluate the sensor measurement from both the left and right light sensor. The larger value is used as the blink value that is stored in the array and can be used in the participant code.
The blink value for each light sensor is influenced by four factors:
\begin{enumerate}
    \item $\Delta$, the distance factor related to the distance $d$ between the sensor and the blink light
    \item $\beta$, the viewing angle factor related to the angle $\alpha$ between the forward sensor direction and the direction from the sensor to the blink light
    \item $\phi$, the relative cube orientation correction dependent on the angle $\varphi$ between the direction from the sensor to the blink light and the direction perpendicular to the edge of a cube's base plate
    \item $\delta$, the relative cube orientation distance correction, the effect mentioned above is dependent on the distance $d$ between the sensor and blink light
\end{enumerate}

The current blink value in the simulation is calculated by:
$$ blink value = \Delta(d)  \beta(\alpha) - \Delta(d)  \beta(\alpha) \cdot \phi(\varphi) \delta(d) $$
The blink value neglecting the cube orientation towards the sensor is $\Delta(d)  \beta(\alpha)$ but the actual blink value can be decreased due to the orientation of the cube. The cube orientation correction $\phi(\varphi)$ is furthermore dependent on the distance and corrected by multiplying $\phi(\varphi)$ by a factor $\delta(d)$. The product of $\phi(\varphi) \cdot \delta(d)$ is a relative correction and therefore must be multiplied with $\Delta(d)  \beta(\alpha)$ to get the absolute correction.


\begin{figure}
  \centering
    \includegraphics[width=0.5\textwidth]{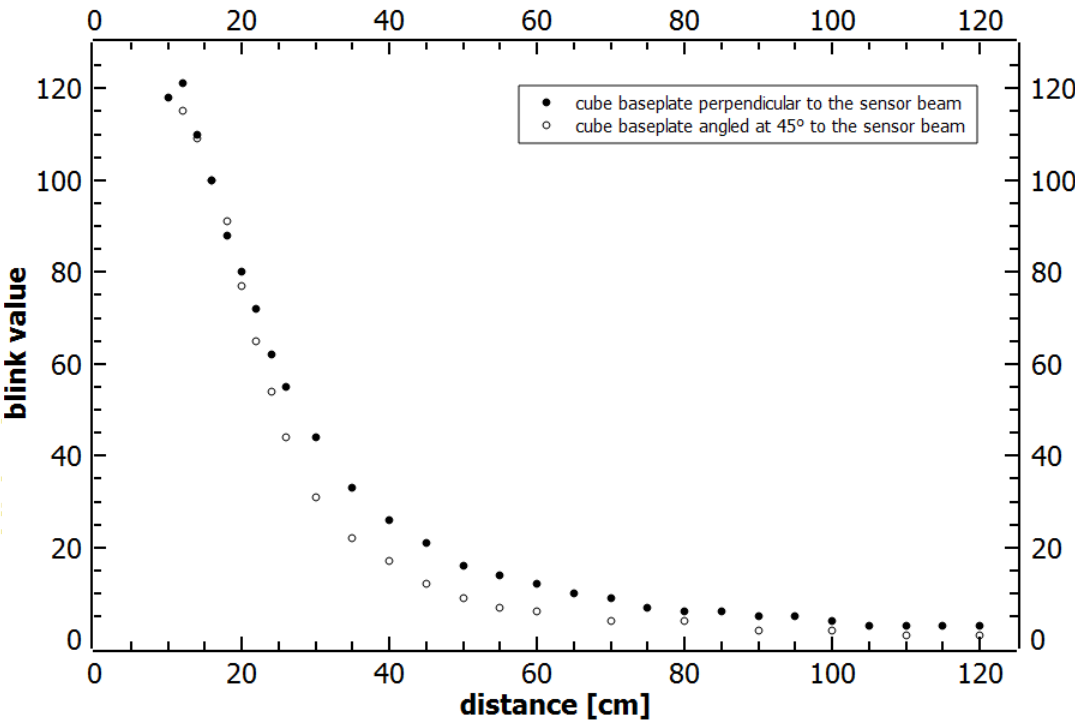}
    \caption{Blink values dependent on the distance from the sensor to the blink light in a cube for the blink light positioned straight ahead of the sensor. The black symbols represent the measurement for a cube orientation where the base plate of the cube is perpendicular to the forward sensor direction. The white symbols represent a cube's base plate oriented at 45 degrees to the sensor.}
    \label{fig:blinkdistance}
\end{figure}

\begin{figure}
  \centering
    \includegraphics[width=0.5\textwidth]{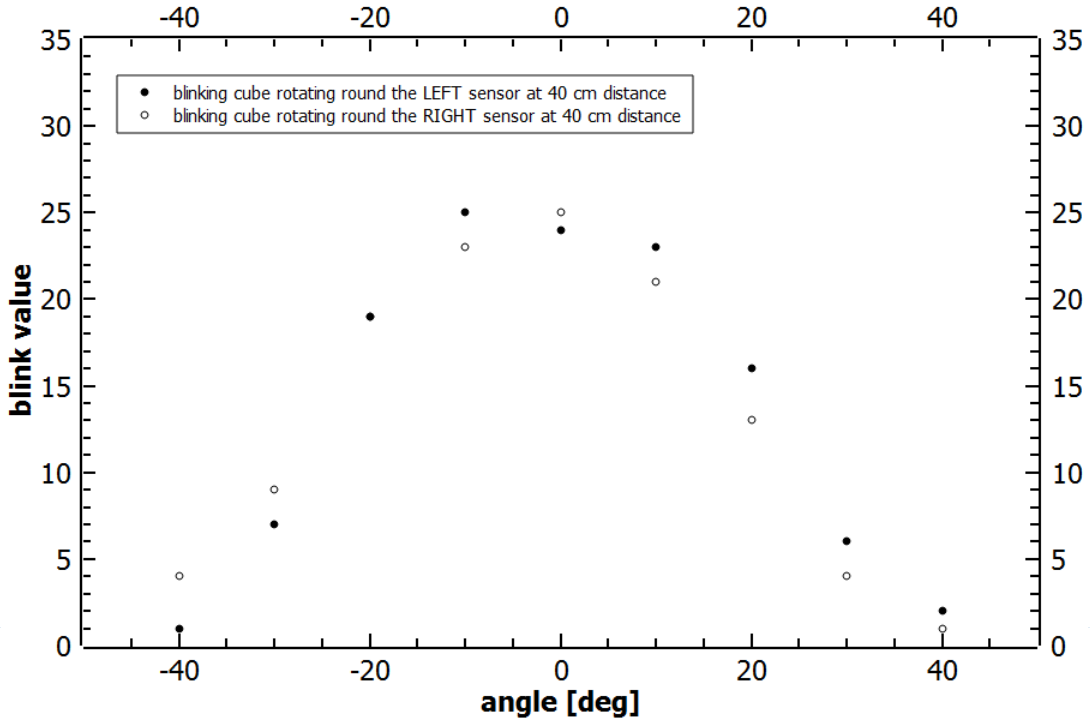}
    \caption{Blink values for the left sensor (black symbol) and the right sensor (white symbol) at a distance of 40 centimeters depending on the angle between the forward direction of the sensor and the direction to the blink light. The angle is negative if the blink light is positioned to the left of straight forward direction from the sensor. At all angles the cube's base plate is held perpendicular to the direction from the sensor to the blink light. }
    \label{fig:viewangle}
\end{figure}

\begin{figure}
  \centering
    \includegraphics[width=0.5\textwidth]{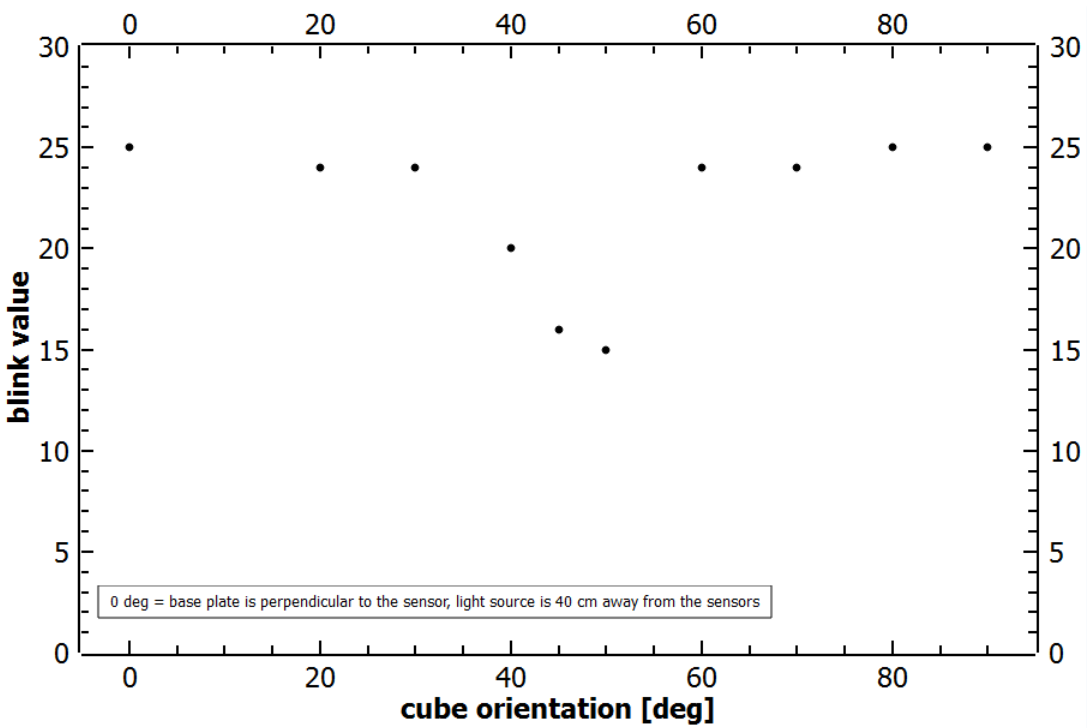}
    \caption{Blink values for a blink light inside a cube positioned 40 centimeters straight in front of the sensor for different orientations of the cube relative to the sensor. Angles of zero and ninety degrees represent a cube with a base plate perpendicular to the forward direction of the sensor. At 45 degrees the corner of the cube is pointing towards the sensor.}
    \label{fig:cubeorientation}
\end{figure}

The distance factor is determined from the measurements shown in figure \ref{fig:blinkdistance}. We use the measurement for the perpendicular orientation of cube's base plate relative to the forward sensor direction. The distance factor $\Delta$ is fitted with

$$ \Delta = 3 \cdot 10^{-6} d^4 - 0.0011 d^3 + 0.144 d^2 - 8.55 d + 200 $$

This blink value based on the distance gets influenced when the blink light is not positioned straight in front of the sensor. The blink value decreases if the blink light is positioned towards the sides of the sensor. This effect is also measured at a sensor to blink light distance of 40 centimeters and normalized to the blink value of 24. The measurements for the different sensors on each side deviate slightly from one another and are shown in figure \ref{fig:viewangle} and the normalized fits are:

$$ \beta_{left} = \frac{1}{24} (8 \cdot 10^{-6} \alpha^4 + 7 \cdot 10^{-5} \alpha^3 - 0.0277 \alpha^2 - 0.1567 \alpha + 25.023)$$
$$ \beta_{right} = \frac{1}{24} (7 \cdot 10^{-6} \alpha^4 + 7 \cdot 10^{-5} \alpha^3 - 0.027 \alpha^2 - 0.0891 \alpha + 26.005)$$

The uncorrected blink value can then already be calculated to $\Delta(d) \cdot \beta(\alpha)$ for each of the sensors. However, if the cube base plate is orientated at angles above 30 and below 60 degrees to the forward direction of the sensor the blink value decreases.
This dependency is shown in figure \ref{fig:cubeorientation} where an angle of zero or ninety degrees is equivalent to the base plate being perpendicular to the sensor direction. The relative correction is fitted from this data to:
$$\phi = \frac{1}{24}(24 - (0.0338 \cdot \varphi^2 - 3.0929 \cdot \varphi + 87.133)$$
The fit is accurate for angles between 30 and 60 degrees. Outside of this interval the simulation assumes no influence caused by the cube orientation. The influence of the cube orientation is distance dependent. This can be seen from the difference of the two curves in figure \ref{fig:blinkdistance}. We use the blink value difference of these two curves and the blink value difference between zero and forty five degrees cube base plate orientation from figure \ref{fig:cubeorientation} to fit a correction $\delta(d)$ of the cube orientation correction.
$$ \delta = 3 \cdot 10^{-9} d^5 - 10^{-6} d^4 + 0.0002 d^3 - 0.0144 d^2 - 0.482d - 4.6 $$

We are aware that these polynomial fits are over-fitted outside of the domain that is relevant for the simulation. This is justified because we want to make a precise interpolation in the domain the sensor values are changing and we can program a constant for where the sensor values are no longer influenced.

\subsubsection{Colour light reflection sensor}
The sensor of the physical robot sends out coloured (red, blue and green) light and measures how much light is reflected back into the sensor. In the experiment and the simulation only the red reflection is used. The reflection effect cannot be estimated by using a single distance between the sensor and an object but is ideally simulated by reflection of infinitely many light rays. We tried and determined that it is possible to simulate the red reflection sensor by using 21 raycasts for each of the sensors equally spaced within the view wangle of the sensors to get a simulated value that is consistent and within the error margin of the physical sensors. The rays originate from the sensor location and measure the distance until they hit an object. The rays are sensitive to a polygon collider in the shape of the crossed walls of the physical cubes and a circular collider. The circular collider is added to decrease the distance from the sensor to the reflecting object because when the simulated sensor is directed at the cube center from a close distance the simulated value was smaller than the actually measured value. We cannot logically explain why this manipulation is necessary but we justify this by wanting to accurately imitate the physical sensor even if we do not understand why the one to one replication of the physical environment does not give the desired results.

\begin{figure}
  \centering
    \includegraphics[width=0.5\textwidth]{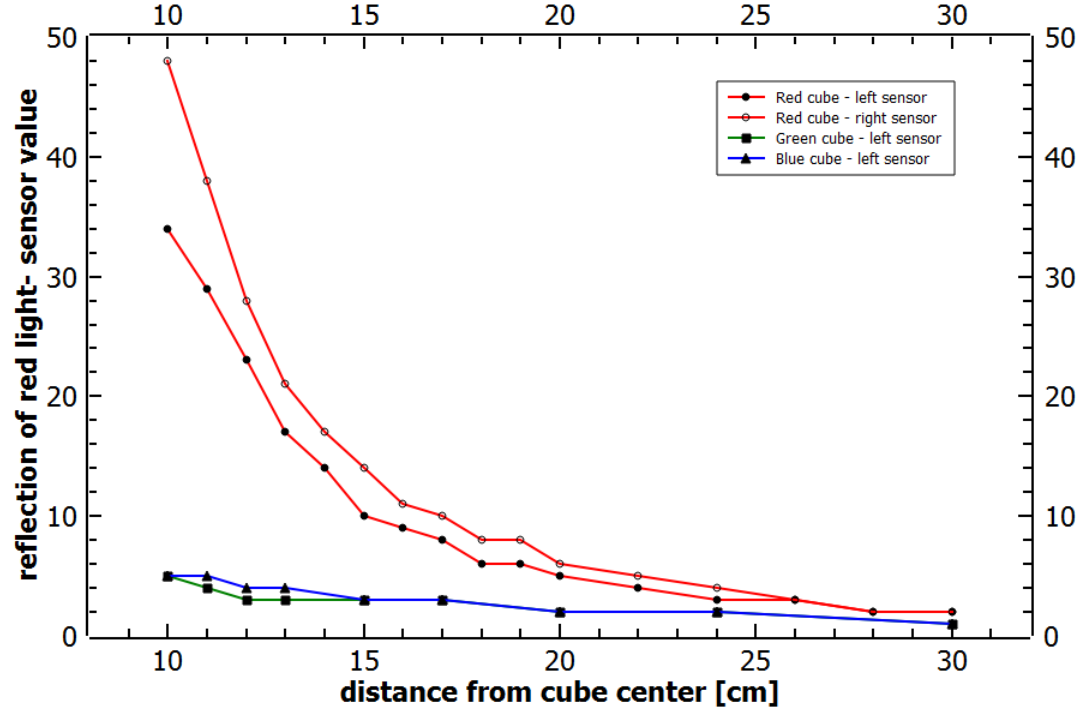}
    \caption{distance dependency of the reflection values of red light for a red, green and blue cube straight in front of the sensor. The distance is measured between the sensor and the cube center.}
    \label{fig:redReflection}
\end{figure}

The simulation is calibrated to the reflectivity measured with the physical robot and cubes.
Figure \ref{fig:redReflection} shows the red reflection values of each sensor for different distances between the sensor and the cube. The sensor is aligned towards the center of the cube. These curves are fitted and the fit formulas are used to estimate the reflectivity for each raycast for each sensor. For each of the sensors the values for all 21 raycasts are then averaged to get the overall value for this sensor. 
The fit function for the reflection of red light from a red cube for the left sensor is:
$$ red_{left} = -0.0114 d^3 + 0.8174 d^2 -19.52 d + 158.94 $$
and the fit for the reflection of red light from a red cube for the right sensor is:
$$ red_{left} = -0.0177 d^3 + 1.244 d^2 - 28.979 d + 229.03 $$.

The measurements for the reflection of red light from green and blue cubes for both sensors are significantly smaller and we consider them similar within the measurement precision of the sensors. Therefore, we estimate them with one similar fit which is:
$$ bluegreen = -0.1572 (d - 5) + 5.5442 $$.

The calibration was tested by placing the physical robot in front of the cubes and validating this setup with the simulated values. Special focus was on when the cube fills the view angle of the sensor only partly, when the sensor points onto an edge of the cube walls and when the cube is locked in the pushing location central in front of the robot.

\section{Verification of the simulation}
\subsection{Why and how we verify}

The goal of the simulation is to have a more repeatable, accurate and time-saving tool for evaluating the performance of the programmed solution with a high number of runs rather than using the physical robot. The simulation can only be used as a substitute for running the physical robot if the simulated robot behaviour is an accurate enough representation of the physical robot's behaviour. 
Thus, the physical robot has to be well characterized in order to be the reference for the development and evaluation of the simulation.
We will describe the three solutions that we use to validate the simulation. We describe how the physical and the simulated robot behave and compare them qualitatively and quantitatively. For the qualitative comparison the behaviour of the physical robot is judged from videos showing the trajectories and interactions with the cubes and the boundaries.
For the quantitative comparison we answer the following questions: 
\begin{enumerate}
    \item How accurately can the physical robot repeat its performance under the same starting condition?
    \item Does the simulation give the same results as the physical robot?
\end{enumerate}

\subsection{The test codes}
Three solutions that reliably solve the task are chosen such that they include all sensors and functions necessary for detecting and removing the cubes.

\subsubsection{ultrasound solution 1} 
This ultrasound based solution drives straight ahead whenever the ultrasound sensor detects an object within a distance of 120 centimeters or otherwise turns clockwise. The robot does not fall off the cardboard because the cubes, that are detected by the ultrasound sensor, are removed as soon as they are fully outside the white area. The physical cubes are removed by the experimenter and the digital cubes are programmed to disappear when the cube is outside the white area.

\subsubsection{ultrasound solution 2} 
The second solution is also ultrasound based. It also turns clockwise until the ultrasound sensor detects  object within a distance of 120 centimeters. Then it turns further to fully face the cube. Then it drives straight forward until it reaches the edge of the white area and continues driving for half a second longer to ensure that the cube is fully pushed outside of the white area. It then turns clockwise back towards the white area and continues searching cubes with the ultrasound sensor. This code does not react when it misses or looses a cube after detecting it.

\subsubsection{blink solution}
This solution uses blink lights in each cube to detect cubes. It turns clockwise in 20 degree steps until the robot measures a blink value that is larger than 13. After each turn the robot stops in this position for 1 second to allow the blink value to update to the correct value for this robot orientation. The robot aborts the search after 18 turning steps (one full rotation) without finding a blink light and then drives straight for two seconds in order to start a new search in a different area. If the robot detects a blinking light it drives straight ahead until it reaches the edge of the white area or it no longer detects a blink light in front of the robot for more than one second.

\subsection{Qualitative results}
When qualitatively comparing the behaviour of the physical robot with the simulated robot we start both scenarios with the same starting configuration and with the same code. We observe the trajectories of the robots and how the physical and simulated trajectory diverge, how the robots interact with the cubes and if they are removed in the same order and generally if there are any repeating patterns such as the robot getting stuck in corners or missing cubes after initially detecting them.

\subsubsection{ultrasound solution 1}
The physical robot turns until it detects a cube and then drives straight forward. If the robot is not directly heading towards the center of the cube it will eventually loose the cube out of the view angle of the ultrasound sensor and start turning again until the cube falls into the view angle again. Depending on the distance between the physical robot and the cube it performs one to three correction turn until it touches the cube. Theoretically we would expect many more corrections but we believe that the angular momentum of the robot when it turns leads to an overshoot so that the cube lies firmly within the view angle of the ultrasound sensor when the robot begins to drive straight forward. The overshoot occurs either because the robot's angular momentum makes it impossible for the robot to change directions abruptly or because a delay in the ultrasound measurement causes a delayed command to the motors to drive straight. We can replicate the overshoot of the physical robot in the simulation by both introducing a delay in the ultrasound measurement and by simulating a limited acceleration. Since the limited acceleration is also observable in the physical robot and we cannot identify a delay of the ultrasound measurement we implemented only the limited acceleration in the simulation. Both the physical and simulated robot remove the cubes in the same order and from the same quadrant of the white area in the three tested starting configurations. When cubes are located near the boundary and the robot is located on the opposite side of the playing field the detection range can be insufficient and the robot keeps turning infinitely. This occurs with the physical robot as well as the simulation.

\subsubsection{ultrasound solution 2} 
The physical robot rotates until the ultrasound sensor detects a cube and then drives straight forward. For cubes more than eighty centimeters away it occurs that the robot does not fully turn towards the cube and misses it and keeps driving straight without corrections until it reaches the white area. In case no correction is needed, i.e. in general with short distances to the cubes this code performs more efficiently than ultrasound solution 1 because it approaches the cubes in a straight line and avoids driving an additional distance. However, we expect that in the average over many trials this solution is inferior because it looses a lot more time when it misses cubes.
The simulated robot generally shows a similar behaviour. It also turns towards the cubes, detects them approximately at equal angles and occasionally misses cubes. However, it misses cubes far more seldom. So far we cannot explain this difference as it does not seem cube distance dependent and predictable.
The different trajectories of the physical robot when restarting the robot under the same starting conditions can differ largely. This is because if the robot misses a cube and drives past it the situation is vastly different and the robot's reaction to the different situation is also very different. Instead of one repeating scenario we observe a range of different scenarios depending on if a cube was removed or not. The simulated trajectory qualitatively matches one of those scenarios. This ultrasound solution has similar issues with the ultrasound detection range both physically and in the simulation.

\subsubsection{blink solution}
The detection range using the blink lights is shorter than the detection range for the ultrasound sensor. The robot therefore often cannot find a cube when turning around and searching. The physical and the simulated robot most of the times detect and remove the same cube first but less reliably as for the previous codes. This means that the trajectories of the repetitions differ significantly depending on if a certain cube was detected or not. The simulation reproduces the scenario that was most common during the tests with the physical robot and is then usually accurate for the first cube and in many cases also the second cube. The simulated trajectory for removing the last cube was never similar to any trajectory obtained using the physical robot.
The robot then continues to drive to a different location and starts searching again. When the cube is near the center it is far more likely to fall into the detection range than when the cube is near the boundary. The blink solution often moves cubes away from the center but does not fully move them outside the boundary. The cubes near the boundary are far less likely to be detected and in these cases the task completion times are significantly longer. This phenomenon occurs equally in the simulation and in reality.

\subsection{Quantitative results}
Ensuring that the simulation matches the behaviour of the physical robot qualitatively is a good starting point. If we want to use the simulation for quantitatively evaluating the performance of solutions we need to validate that simulation estimates the quantitiave result of the physical robot accurately.
For this we start with measuring the repeatability of the physical robot by repeatedly executing the same scenario. We then execute the physical robot and the simulation from many different conditions and compare the results statistically.

\subsubsection{How repeatable is the performance of the physical robot?}
The three described solutions are evaluated with the physical robot from three different starting positions with up to 13 repetitions from each starting position. We show the mean time the robot needed to complete the task and the standard deviation over the $N$ repetitions. Failed attempts are excluded from the statistics as they would imply an infinite task completion time. The failed attempts are caused by the robot getting stuck outside the cube detection range.

\begin{table}[!htb]
\centering
\caption{Results of repeatedly executing three solutions with the physical robot from three different starting conditions}\label{tab1}
\begin{tabular}{lccccc}
\hline\\[-1.5ex]
& start pos. & $M$ & $SD$ & $N$ & failed \\[0.5ex]
\hline\\[-1.5ex]
ultrasound 1 & 1 & 19.0 & 6.7 & 11 & 1 \\[0.5ex]
 & 2 & 22.6 & 4.3 & 12 & 0 \\[0.5ex]
 & 3 & 26.4 & 9.2 & 13 & 0 \\[0.5ex]
\hline\\[-1.5ex]
ultrasound 2 & 1 & 27.9 & 9.8 & 9 & 1 \\[0.5ex]
 & 2 & 26.6 & 5.7 & 10 & 0 \\[0.5ex]
 & 3 & 30.2 & 0.6 & 10 & 0 \\[0.5ex]
\hline\\[-1.5ex]
blink solution & 1 & 89.9 & 45.8 & 7 & 0 \\[0.5ex]
& 2 & 88.4 & 53.3 & 8 & 0 \\[0.5ex]
& 3 & 110.8 & 98.5 & 8 & 0 \\[0.5ex]
\hline
\end{tabular}
\end{table}

\subsubsection{Statistical comparison of the physical vs. the simulated robot}

We measure the task completion time for the three solutions from twenty different starting configurations using the physical robot and the simulated robot. The starting configurations used in the physical evaluation and the simulation are equal and each starting position is run once. We present the mean task completion time and its standard deviation for each solution over those twenty runs. $N$ describes the number of included measurements. If $N$ is less than twenty this means that the solution failed $20 - N$ times. This is the case for the ultrasound based solutions and this happened when the nearest cube object was outside the ultrasound detection range when the robot was searching for cubes. These runs were aborted because they would lead to an infinite task completion time. Failed attempts are excluded from calculating the mean and standard deviation.

\begin{table}[!htb]
\centering
\caption{Comparison of the completion times, physical vs. simulated robot for 20 starting positions}\label{tab4}
\begin{tabular}{llcccc}
\hline\\[-1.5ex]
 & solution & $N$ & $M$ & $SD$ & \% of phys. $M$\\[0.5ex]
\hline\\[-1.5ex]
physical & ultrasound 1 & 18 & 20.5 & 2.92 & 100\\[0.5ex]
 & ultrasound 2 & 19 & 28.6 & 11.06 & 100 \\[0.5ex]
 & blink & 20 & 106.6 & 43.07 & 100\\[0.5ex]
\hline\\[-1.5ex]
simulated & ultrasound 1 & 20 & 15.25 & 1.38 & 74.4 \\[0.5ex]
 & ultrasound 2 & 19 & 20.28 & 3.54 & 70.4 \\[0.5ex]
 & blink & 20 & 84.21 & 72.58 & 78.4 \\[0.5ex]
\hline
\end{tabular}
\end{table}

For the initial placement of the robot for the twenty starting positions the robot's front edge is aligned along rectangular lines on the white area. This means that the initial rotation is along one of the four major directions and the locations are incremental.
This setup is chosen such that the initial placement of the robot is easily repeatable. To check if we introduce a bias by selecting the starting configurations in this way we run the simulation from 99 randomly chosen starting positions and orientations. The random starting configurations are generated with a random number generator in the Unity software. Starting configurations where the robot is initially touching a cube are removed. The results are shown in table \ref{tab6}:

\begin{table}[!htb]
\centering
\caption{Simulated task completion times for 99 randomly assigned, non-rectangular starting positions and orientations}\label{tab6}
\begin{tabular}{lccc}
\hline\\[-1.5ex]
solution & $N$ & mean & std.dev. \\[0.5ex]
\hline\\[-1.5ex]
ultrasound 1 & 99 & 15.16 & 1.06\\[0.5ex]
ultrasound 2 & 99 & 20.86 & 7.37\\[0.5ex]
blink & 99 & 75.53 & 57.66\\[0.5ex]
\hline
\end{tabular}
\end{table}

In the discussion we will argue that the measured task completion time is sensitive to perturbations. To investigate this we run the three codes from the previously used twenty starting positions and initially orient the robot one degree further counter-clockwise.
We calculate the pairwise differences in task completion time between the non-perturbed and the perturbed starting orientation for each of the twenty starting configuration and then take the mean and standard deviation of the differences. The results are shown in table \ref{tab7}.

\begin{table}[!htb]
\centering
\caption{Influence of one degree counter clockwise perturbed starting orientation on task completion time}\label{tab7}
\begin{tabular}{lccccc}
\hline\\[-1.5ex]
 &  & pert. & pert. & diff. & diff. \\
solution & $N$ & mean & std.dev. & mean & std.dev. \\[0.5ex]
\hline\\[-1.5ex]
ultrasound 1 & 20 & 14.97 & 1.85 & 0.35 & 0.67 \\[0.5ex]
ultrasound 2 & 19 & 19.05 & 3.28 & 1.26 & 1.07 \\[0.5ex]
blink & 20 & 64.00 & 32.20 & 42.57 & 59.32 \\[0.5ex]
\hline
\end{tabular}
\end{table}

Another imperfection of the physical robot is that the motors do not perfectly keep the rotation speeds that they are assigned in the code. To estimate the influence of inaccurate motor speed we simulate the three solutions from the previously used twenty starting positions with a one percent motor speed increase on the right motor. This means that the robot drives a visually unnoticeable left turn when it is originally programmed to drive straight. 

\begin{table}[!htb]
\centering
\caption{Influence on 1\% faster motor speed of the right motor on task completion time}\label{tab8}
\begin{tabular}{lccccc}
\hline\\[-1.5ex]
 &  & pert. & pert. & diff. & diff. \\
solution & $N$ & mean & std.dev. & mean & std.dev. \\[0.5ex]
\hline\\[-1.5ex]
ultrasound 1 & 20 & 14.89 & 2.1 & 1.07 & 1.03 \\[0.5ex]
ultrasound 2 & 19 & 20.00 & 4.34 & 2.23 & 2.51 \\[0.5ex]
blink & 20 & 62.10 & 28.20 & 39.97 & 65.06 \\[0.5ex]
\hline
\end{tabular}
\end{table}

\subsection{Discussion of the results}
We want to use the simulation to compare the performances of code solutions written for the physical robot in a more time saving and more controlled way. The simulation needs to reproduce the results obtained with the physical robot at least relatively to compare different solutions. Most intuitively one tries to program a simulation that matches the reality as accurately as possible by comparing and improving the simulation to match reference cases taken with the physical robot. This entails that the physical and the simulated robot are deterministic and thus reliably repeat their respective behaviours when started under the same conditions. The simulation is deterministic but the physical robot shows significant deviations both in its trajectory as well as in the task completion time which we use as the main performance metric. Table \ref{tab1} gives the task completion times of the physical robot for three different solutions from three different starting positions of the robot. We see that the deviations from the mean depend primarily on the solution and also on the starting position. For the ultrasound solution 1 the standard deviation of the task completion time is between 19\% and 35\% of the mean, for ultrasound solution 2 it is between 1,4\% to 35\% and for the blink solution it is between 50\% and 89\%. Apart from the one starting position with the low standard deviation for ultrasound solution 2 the deviations are too large to use the trajectories and task completion times as references for developing a simulation. As a consequence, despite the simulation being in itself accurately repeatable, it cannot evaluate a solution more accurately than how accurately the physical robot can repeat its performance, because we have no clear reference to optimize the simulation to. This means that a the simulation from one starting configuration, despite being numerically precise, has the same uncertainty as a single evaluation of this starting configuration with the physical robot. 
We develop the simulation by qualitatively observing the physical robot and the simulation and adjust the simulation until it qualitative shows the same characteristic behaviour, such as interactions with cubes and the boundary, in as many cases as possible. This approach aims to make the simulated behaviour qualitatively indistinguishable from the physical robot behaviour. However, this qualitative validation of the simulation is subjective and depends on the judgment of the experimenter looking at the video captures of the physical robot.

The large deviations of the physical robot originate from small perturbations that can neither be controlled for in physical measurements nor can the simulation replicate reality with more precision than the perturbations that cause the deviations. Therefore, a single measurement of the physical or the simulated robot is not a good indication for the overall performance of the solution. However, we assume that these perturbations are normally distributed and their effects can be mitigated by repeatedly executing and evaluating the robot from many starting conditions and performing statistics. To illustrate this approach we measure and simulate the task completion times for the three solutions from twenty different starting positions. The measured and the simulated task completion times seldom match pairwise for a single starting position but the question is if the results match relatively when comparing the means of the different solutions. The results are shown in table \ref{tab4}. Firstly we see that the means of the simulated task completion time are between 70\% and 78\% of the measured means task completion times, i.e. the simulation does not absolutely give the same results as the measurement with the physical robot. For comparing the performance of different solutions it is sufficient that the simulation gives the same result when relatively comparing the task completion times. We compare the three solutions by normalizing the means and standard deviations by the task completion time of the blink solution, which was the slowest of the three solutions. The means and standard deviations measured with the physical robot are normalized to the physically measured blink solution mean and the simulated results are normalized to the simulated blink solution result. The normalized task completion time mean for the physically measured ultrasound solution 1 is 19,3\% of the blink solution task completion time with a standard deviation of 2,7\% whereas this comparison for the simulated case gives a normalized mean of 18,2\% with a normalized standard deviation of 1,6\%. These normalized means are well within each others standard deviations and are considered equal within the measurement accuracy. The same is the case for ultrasound solution 2 with a normalized mean of 26,9\% and a normalized standard deviation of 10,4\% for the physical evaluation and a normalized mean of 24,1\% and a normalized standard deviation of 4,2\% for the simulated evaluation. The same conclusion is true when normalizing to other values. This means that we can compare different solutions relatively without needing a simulation that is absolutely accurate.
The twenty starting positions for the robot that were used for this evaluation were chosen such that they can be reliably reproduced in the real world and replicated as similar as possible in the simulation. This means that the starting positions were reasonably chosen to cover a wide range of different starting positions and orientations. It can be that the selection caused a bias in the setup. To ensure that this is not the case we simulated the three solutions from 99 randomly chosen starting positions. The results are in table \ref{tab6}. The task completion time means for the simulation with 20 chosen and 99 random starting positions differ by less than 1\% for ultrasound solution 1, by less than 3\% for ultrasound solution 2 and by about 10\% for the blink solution. The increase in relative difference can be explained with the accompanied increase of the standard deviation of these solutions. This shows the importance of a large sample size for solutions with a large standard deviation. The differences between the simulations are significantly less than their standard deviations (and standard errors) and we conclude that the chosen twenty starting positions do not show a significant bias.

During the observation of the physical robot we saw that despite starting the same code and placing the cubes and the robot as similarly as possible at the same positions for each repetition we see that the robot initially drives very similar trajectories that diverge from each other over time and some events cause a drastically different trajectory afterwards. These drastic differences between trajectories occur when one trajectory removes a cube while the other only moves or rotates the cube without removing it or even misses the cube entirely. Another scenario is when the robot is programmed to react when it reaches and detects the boundary of the white area and the trajectories are almost aligned with the boundary. An initially small perturbation in the location or orientation of the robot then has a large influence on where it detects and reacts to the boundary causing a larger deviation of the trajectories. Both scenarios make the system sensitive to small differences in robot location and orientation. To illustrate this sensitivity and investigate the effect of a single perturbation under controlled conditions we use the previous twenty starting conditions to simulate the task completion times when the initial robot starting orientation is changed by 1 degree counter clockwise. The mean task completion time from the unperturbed starting positions in table \ref{tab4} and the mean task completion time from the counter clockwise rotated starting positions in table \ref{tab7} are similar within their uncertainty limits. We cannot distinguish a significant effect on the mean task completion times caused by perturbing the starting positions. We then calculate the absolute pairwise difference between the task completion times of the unperturbed and perturbed simulation for each of the twenty starting positions. The mean of these absolute differences and the standard deviation thereof is also shown in table \ref{tab7}. We see that the misalignment of the robot by one degree already explains $ 0.35 / 1.38 = $ 25\% of the standard deviation observed over the twenty different unperturbed starting positions for ultrasound solution 1. For ultrasound solution 2 the misalignment accounts for 36\% of the original standard deviation and for the blink solution the perturbation causes an absolute difference mean that is 59\% of the standard deviation. We repeated the same idea of comparing perturbed and unperturbed robot behaviour with increasing the motor speed of the right motor by 1\% point. The results are shown in table \ref{tab8} and the mean task completion time is similar to the unperturbed result within the measurement accuracy. From the pairwise differences in task completion time for each starting position and its mean we see that it explains 78\% of the unperturbed standard deviation of ultrasound solution 1, 63\% for ultrasound solution 2 and 55\% for the blink solution.
Both perturbations have a significant influence on the task completion time for a single evaluation from a single starting position. These perturbation are very likely to occur when executing the physical robot and are both undetectable and therefore impossible to control for. This explains why we observe significantly varying results when repeating the physical measurement under conditions that appear similar but are in fact perturbed. This makes it impossible to replicate the physical conditions in the simulation with a precision that has a negligible effect on the trajectory. Evaluating the performance of different solutions from a single measurement is not possible. Evaluating the performance requires repeated measurements and statistics. Manually executing and measuring the physical robot often enough to get a reliable statistic result can in principle be done but is very time consuming and therefore not feasible. Additionally, can the simulation keep the conditions absolute repeatable whereas an evaluation with the physical robot is subject to drainage of the batteries, wear and tear of the mechanical parts and changes of the surrounding conditions such as light and temperature that cause perturbations that have significant influences on the repeatably as illustrated above. These influences make a comparison of different solutions under similar conditions impossible and the inaccuracy introduced by not perfectly matching the behaviour of the physical robot within the simulation is far less significant than the inaccuracy introduced by changing conditions. Another advantage of the simulation is that it is very time saving, allows to generate more results from more starting positions and therefore allows a statistical analysis with a lower uncertainty.

\subsection{Conclusion}
We develop a simulation to a self-built LEGO Mindstorms robot. We aim to use the simulation to compare the performance of different code solutions programmed to solve a task that involves finding and moving cubic objects. We replicate the physical environment digitally and measure the behaviour of the physical robot to make a mathematical model that we use to implement a simulation. The simulation is primarily based on these measurements and then adjusted from qualitative comparisons between the physical and the simulated robot behaviour.
After adjusting and verifying the simulation for the individual sensors and actuators we verify the simulation by comparing the simulated robot behaviour with the behaviour of the physical robot in a systems approach with three solutions that cover the entire scope of available functions combined in solutions that solve the task. We conclude that the trajectories of the physical robot deviate significantly despite being repeatedly executed under as similar as possible conditions as possible. The physical robot behaviour is not deterministic. This implies that the simulation cannot accurately replicate the trajectory of the physical robot when the physical robot does not have a consistent trajectory. We verify that the simulation qualitatively shows a similar behavioural characteristic as the physical robot. For quantitatively comparing code performances a repeated evaluation from different starting conditions and a statistical analysis is required. We compare the task completion times of the physical and the simulated robot for twenty different starting conditions. We show that the simulation does not absolute give the same result as measured physically but it is relatively accurately when comparing different solutions. We ensured that the starting conditions used for the quantitative verification are unbiased.
We argue that the sometimes drastically different trajectories and task completion times of the physical robot are caused by a sensitivity of the system to small perturbations of the robots position and orientation. We use the simulation to demonstrate that likely inaccuracies such as a change of the robot's initial orientation by one degree or a one percent speed difference in the motor speed can already explain
the standard deviation calculated from the task completion times of the twenty starting conditions.
Overall we conclude from these results that this simulation is needed and a useful tool for relatively comparing solution performances under controlled conditions with sufficient data for statistical methods.   

\bibliographystyle{unsrt}  
\bibliography{ms}

\begin{thebibliography}{1}

\bibitem{Gerstenberg_2018_a}
Achim Gerstenberg and Martin Steinert.
\newblock Open ended problems - a robot programming experiment design to
  compare and test different development and design approaches.
\newblock In {\em DS 91: Proceedings of NordDesign}. Design Society, 2018.

\bibitem{datasheet}
Achim Gerstenberg.
\newblock robotdatasheet.pdf in robotexpsetup on github, May 2018.
\newblock \url
  {https://github.com/AchimGerstenberg/RoboExpSetup/blob/master/printouts/robotdatasheet.pdf}
  (accessed 24-March-2019).

\bibitem{unity}
Unity~Technologies ApS.
\newblock Unity version 5.6.1f1, May 2017.
\newblock \url {https://unity.com/} (accessed 24-March-2019).

\end{thebibliography}

\end{document}